\documentclass[conference]{IEEEtran}
\usepackage{cite}
\usepackage{amsmath,amssymb,amsfonts}
\usepackage{algorithmic}
\usepackage{graphicx}
\usepackage{textcomp}
\usepackage{xcolor}

\usepackage{array}
\usepackage{multicol}
\usepackage{multirow}

\usepackage{cleveref}
\crefformat{section}{\S#2#1#3} 
\crefformat{subsection}{\S#2#1#3}
\crefformat{subsubsection}{\S#2#1#3}
\usepackage{blindtext}

\usepackage{cite}
\usepackage{amsmath,amssymb,amsfonts}
\usepackage{eufrak}
\usepackage{algorithmic}
\usepackage{graphicx}
\usepackage{subfigure}
\usepackage{textcomp}
\usepackage{xcolor}
\usepackage{comment}

\usepackage[letterpaper, left=0.625in, right=0.625in, top=0.75in, bottom=1in]{geometry}

\IEEEoverridecommandlockouts

\newcommand{\hytt}[1]{\texttt{\hyphenchar\font=\defaulthyphenchar #1}}

\def\BibTeX{{\rm B\kern-.05em{\sc i\kern-.025em b}\kern-.08em
    T\kern-.1667em\lower.7ex\hbox{E}\kern-.125emX}}

\makeatletter

\def\@fnsymbol#1{\ensuremath{\ifcase#1\or *\or \dagger\or \ddagger\or
   \mathsection\or \mathparagraph\or \|\or **\or \dagger\dagger
   \or \ddagger\ddagger \else\@ctrerr\fi}}

\begin{document}

\title{{\fontsize{18}{0}\selectfont EPAM: A Predictive Energy Model for Mobile AI\vspace{-0.2 in}\\}

}


\author{
\IEEEauthorblockN{Anik Mallik\IEEEauthorrefmark{1}, Haoxin Wang\IEEEauthorrefmark{2}, Jiang Xie\IEEEauthorrefmark{1}, Dawei Chen\IEEEauthorrefmark{3}, and Kyungtae Han\IEEEauthorrefmark{3}\\
\IEEEauthorblockA{\IEEEauthorrefmark{1}\textit{Department of Electrical and Computer Engineering, The University of North Carolina at Charlotte, NC, USA}}
\IEEEauthorblockA{\IEEEauthorrefmark{2}\textit{Department of Computer Science, Georgia State University, GA, USA}}
\IEEEauthorblockA{\IEEEauthorrefmark{3}\textit{InfoTech Labs, Toyota Motor North America R\&D, Mountain View, CA, USA}}\vspace{-0.7 in}}
\thanks{\vspace{-0.01 in}This work was supported by funds from Toyota Motor North America and by the US National Science Foundation (NSF) under Grant No. 1910667, 1910891, and 2025284.}
}

\maketitle

\begin{abstract}
Artificial intelligence (AI) has enabled a new paradigm of smart applications -- changing our way of living entirely. Many of these AI-enabled applications have very stringent latency requirements, especially for applications on mobile devices (e.g., smartphones, wearable devices, and vehicles). Hence, smaller and quantized deep neural network (DNN) models are developed for mobile devices, which provide faster and more energy-efficient computation for mobile AI applications. However, how AI models consume energy in a mobile device is still unexplored. Predicting the energy consumption of these models, along with their different applications, such as vision and non-vision, requires a thorough investigation of their behavior using various processing sources. In this paper, we introduce a comprehensive study of mobile AI applications considering different DNN models and processing sources, focusing on computational resource utilization, delay, and energy consumption. We measure the latency, energy consumption, and memory usage of all the models using four processing sources through extensive experiments. We explain the challenges in such investigations and how we propose to overcome them. Our study highlights important insights, such as how mobile AI behaves in different applications (vision and non-vision) using CPU, GPU, and NNAPI. Finally, we propose a novel Gaussian process regression-based general predictive energy model based on DNN structures, computation resources, and processors, which can predict the energy for each complete application cycle irrespective of device configuration and application. This study provides crucial facts and an energy prediction mechanism to the AI research community to help bring energy efficiency to mobile AI applications.
\end{abstract}

\vspace{-0.15in}

\vspace{-0.15in}
\section{Introduction}
\vspace{-0.05 in}

Artificial intelligence (AI) is shaping every aspect of human lives nowadays. Furthermore, mobile devices, i.e., smartphones, tablets, wearable devices, and autonomous and unmanned aerial vehicles, are heavily invested in AI applications, having cellular networks, edge, and cloud computing in the backbone. AI applications consume considerably high energy and memory of these devices. How AI uses these resources defines a device's potential to interact with wireless networks. Therefore, it is crucial to understand the characteristics of AI applications running on a mobile device, which pushes back to the question --- how can we accurately predict the energy consumption of mobile AI irrespective of device configurations to ensure better service and user experience?




%

\par AI applications' energy consumption may depend on various properties of a system. First, the AI models that are crafted in specific ways to fit mobile devices due to the models' high computation and energy requirements, impact the applications' behaviors. Research works suggest accelerating the processing time of deep neural networks (DNNs) by quantizing \cite{wu2016quantized}, which is a compression technique run on DNN models that can reduce the model size by converting some tensor operations to integers from floating points or reducing the weights or parameters in a model, but at the cost of degraded accuracy. Quantized DNN (Q-DNN) models are generally investigated for vision-based applications, the most thriving areas of AI. Second, mobile AI is not limited to vision applications only. Modern-day mobile devices are rigged with non-vision applications as well, such as intelligent recommendations, natural language processing (NLP), smart reply, speech recognition, and speech-to-text conversion. While most of the research focuses on applications based on computer vision, acquiring a thorough knowledge of mobile AI is only possible by including non-vision applications. Third, the processing source used to run the AI models affects their performance. Besides central processing units (CPUs) with high processing speeds, some devices are now equipped with graphics processing units (GPUs), which enables DNN models to run faster than ever, especially for vision applications \cite{huynh2016deepsense}. In addition, neural network application programming interfaces (NNAPI) are also developed to make the processing of DNN models faster using CPUs, GPUs, or neural processing units (NPUs) \cite{lai2020enabling}. These state-of-the-art technologies are researched for mobile AI only to improve inference latency. Lastly, the hardware configuration of mobile devices is distinctive and contributes to energy consumption with a unique signature. The system-on-chip (SoC), CPU/GPU parameters, and memory dictate how an AI application runs on a specific device.


\vspace{-0.08in}
\par In this paper, we argue that a predictive energy model for a mobile AI application requires considering all of the parameters mentioned above. Without collecting accurate and precise latency, energy, and memory consumption data, it is not possible to design a predictive energy model which is applicable to all AI applications with different model sizes and device configurations. This paper presents the measurement data of AI applications collected through experiments and proposes a novel model of \underline{E}nergy \underline{P}rediction for \underline{A}I in \underline{M}obile devices (EPAM), which can provide a highly accurate prediction of the energy consumption of a mobile AI application irrespective of device configuration and AI models, and thus contribute to improving the overall performance.

\begin{figure}[t!]
\centering
\subfigure[]
{\includegraphics[width=0.226\textwidth]{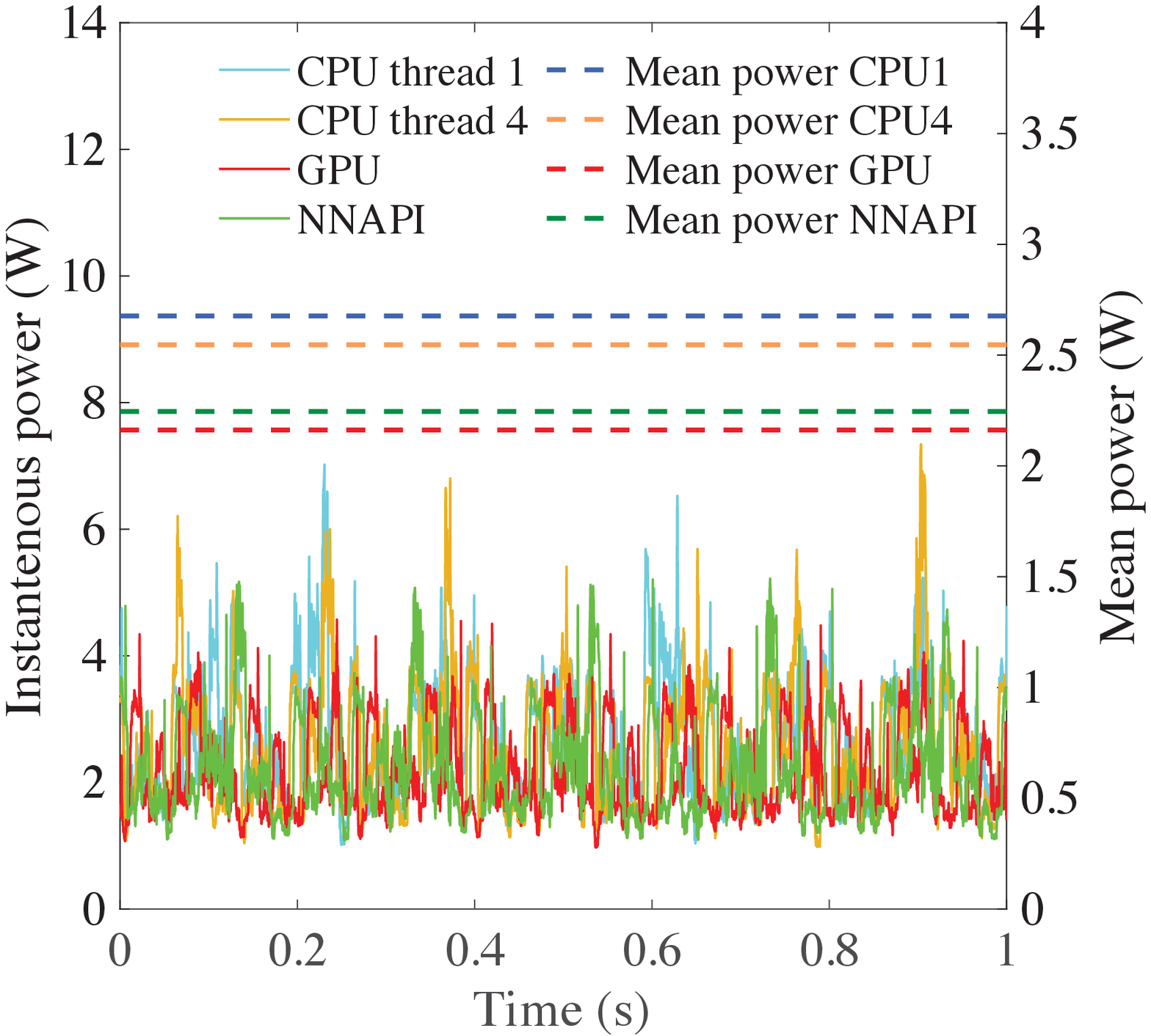}\label{fig: Processors_Motiv}}
\subfigure[]
{\includegraphics[width=0.232\textwidth]{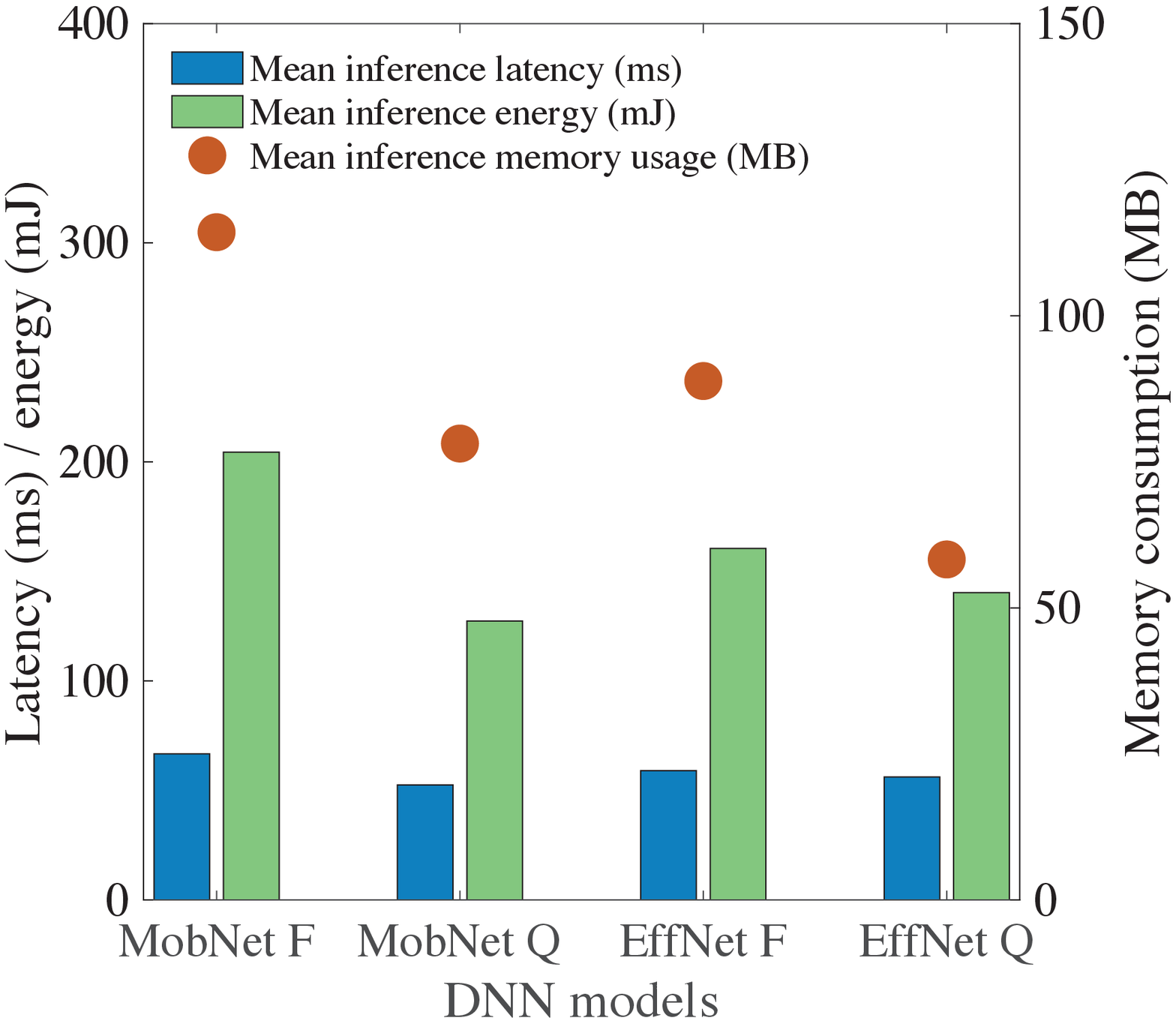}\label{fig: FloatQuant_Motiv}}
\vspace{-0.15 in}
\caption{(a) Power consumption by different processors for the same time interval for MobileNet Float and (b) mean inference latency, energy, and memory usage for float and quantized DNN models on Huawei Mate40Pro.}
\label{fig: Figure_motivation}
\vspace{-0.28 in}
\end{figure}

\vspace{-0.08in}
\textbf{Motivations:} While mobile AI is often concluded as ``no one-size-fits-all solution'' \cite{luo2020comparison}, it is the responsibility of the research community to provide the developers with precise measurement data and a way to predict energy consumption. Our research shows that the power varies for the same device with the change in processing sources (Fig. \ref{fig: Processors_Motiv}). The granularity of power consumption over a unit period of time needs to be measured to develop a predictive energy model, which is not provided by the current works. Battery profilers provided by third-party applications do not support precise energy data collection \cite{wang2021energy}. Hence, the use of an external power measurement tool becomes necessary \cite{mallik2022H264}. Moreover, DNN models with different sizes and layers do not have a similar impact on the latency, energy, and memory usage, which is presented in Fig. \ref{fig: FloatQuant_Motiv}, where it is evident that the correlation among latency, energy, and memory is not linear at all. An interesting observation here is that the Quantized EfficientNet model causes high latency and energy despite using the lowest memory, due to its compatibility issues with NNAPI, which is described in detail in section V-A. This motivates us to collect data from a physical testbed to validate this correlation before proposing a predictive energy model.




\textbf{Challenges:} Designing a predictive energy consumption model for mobile AI is not straightforward. First, a general energy prediction model is challenging to develop \textit{due to different categorical and numerical variables involved in the non-parametric behavior of the energy consumption of AI applications.} The regression model cannot be linear since all the parameters do not have the same weight in all applications and configurations. Second, measuring mobile AI parameters is challenging due to \textit{complicated power terminal design in the latest mobile devices}. Synchronizing the timestamps of latency and energy data brings further difficulties as the retrieved log files have different formats. However, these parameters must be measured since they are required for training the regression model. Finally, the \textit{experiments should be controllable and repeatable} for enthusiastic researchers. Therefore, the environment must be chosen wisely so that all the experiments can be carried out in a similar condition.


\textbf{Our contributions:} Our contributions in this paper are summarized as follows:
\vspace{-0.05 in}
\begin{itemize}
    \item \textbf{Experimental research and analysis of different mobile AI applications:} We set up an experimental testbed with four different smartphones (Table \ref{table:phone specs}) and use a vision application (image classification) and two non-vision applications (NLP and speech recognition) with seven different DNN models (Table \ref{table:DNN}). The testbed is described in detail in Section IV. We investigate different mobile AI parameters through an extensive experimental study. The latency, power consumption, and memory usage of individual segments of the pipelines of three AI applications are measured for different applications using single- and multi-threads CPU, GPU, and NNAPI and for different DNN models. Our experiment shows that the total energy consumption of a mobile AI application is related to the device configuration, AI model, latency, and memory.
    
    \item \textbf{Predictive energy model for mobile AI:} We propose a novel Gaussian process regression-based general predictive energy model for mobile AI (EPAM) based on DNN structure, memory usage, and processing sources to predict the energy consumption of mobile AI applications irrespective of device configurations (Section III). EPAM requires offline training with past datasets. The trained model can be used to predict the overall energy consumption which reduces the necessity for further energy measurement and helps the developers design energy-efficient mobile AI applications. Finally, we evaluate the performance of our proposed predictive energy model EPAM with our experimental data (Section V-D). The evaluation shows that EPAM provides highly accurate energy prediction of vision and non-vision AI applications for different DNN models on unique mobile devices.
    
    
    
    
    
\end{itemize}
\vspace{-0.2in}
\section{Related Work}
\vspace{-0.1in}
\textbf{Vision and non-vision mobile AI with float and quantized models:}
Floating point and quantized models are investigated for vision applications, e.g., image classification, segmentation, super-resolution, and object detection, to create benchmarks using inference latency for mobile devices \cite{ignatov2019ai}. Quantized models are introduced in \cite{cheng2017quantized} to lower the energy consumption as well. In addition, non-vision AI applications are also researched to achieve high accuracy and low latency \cite{iandola2020squeezebert}. Nevertheless, a predictive energy model for mobile AI requires analysis of complete behaviors of vision and non-vision mobile AI applications using floating point and quantized models, which are not yet explored.


\par \textbf{Latency and energy in different processors:}
Mobile AI applications behave differently in terms of latency and accuracy based on the processing sources \cite{ignatov2019ai, luo2020comparison}. Research works are done on maximizing CPU threads \cite{liu2019performance} and hardware acceleration for DNN models. The use of GPU is also studied for improving the training and inference time for mobile AI \cite{huynh2016deepsense}. NPU architectures are explored as well to expedite neural network operations \cite{shin2018dnpu, lai2020enabling}. However, there is no fundamental framework to describe the impact of individual processing sources on energy consumption for different mobile AI applications with disparate DNN models.

\par \textbf{Energy modeling for mobile AI and prediction:} 
Energy measurement is necessary to describe mobile AI applications' detailed behaviors. Eprof \cite{pathak2012energy} and E-Tester \cite{jindal2021experience} are proposed to measure and test the battery drain of mobile devices, which use a finite state machine to measure the energy. However, these methods lack in providing granular and precise energy data since they only act on system call traces. Researchers have proposed different energy models for vision \cite{wang2022leaf} and non-vision \cite{cao2020towards} applications. Furthermore, predictive energy models are developed for devices, and sensors \cite{ullah2022approximating}. Nonetheless, developing accurate predictive energy models general to all mobile AI applications requires knowledge of all the environmental parameters such as network and model size, memory usage, and the hardware accessed to run the AI application.


\section{EPAM: Overview of the Predictive Model} \label{EPAM}
\vspace{-0.05in}
The energy prediction of mobile AI involves a high dimension of influencing variables, making it a non-parametric model. Let us assume that the set of input data points is $X^{1:D}$, where D is the total number of dimensions. If we consider this a noisy observation, then we find the posterior distribution as
\vspace{-0.05in}
\begin{small}
\begin{equation}
P(E(X) \propto P(E(X)|\Lambda^{1:D})/P(\Lambda^{1:D}|E(X)),
\label{eq:1}
\vspace{-0.08in}
\end{equation}
\end{small}\noindent
where $E(X)$ is the observed energy at data points $X^{1:D}$ and $\Lambda^{1:D} = \{X^{1:D}, E\}$ is observation points. Using Gaussian process \cite{wang2020intuitive}, $E(X)$ can be described as $E(X) \sim \displaystyle \mathop{\mathcal{N}}(\mu,K)$, where $\mu = [mean(X^1), \ldots, mean(X^D)]$ is the mean and $K_{ij} = k(x_i, x_j)$ is the covariance or Kernel function, where $x_i$ and $x_j$ are distinct data points. 
\par As new data points $X_*$ are provided, the posterior distribution of predicted energy $E(X_*)$ can be modeled as
\vspace{-0.05in}
\begin{small}
\begin{equation}
P(E(X_*)|\Lambda^{1:D}) \sim \displaystyle \mathop{\mathcal{N}}(\mu(X_*),K(X_*))
\label{eq:2}
\vspace{-0.05in}
\end{equation}
\end{small}

\par \textit{The kernel must be chosen carefully} as there exists a clear link between kernel functions and predictions \cite{seeger2004gaussian}, which contribute to the hyper-parameter optimization. From our experimental data, we observe \textit{the influencing parameters on total energy consumption are sparse and vary over a broad range including both numerical and categorical variables}. Hence, we choose the automatic relevance determination (ARD) exponential squared kernel for our predictive model, which automatically puts different weights on the parameters with differential scales assessing their significance to the model. Hence, our kernel equation becomes:
\vspace{-0.08in}
\begin{small}
\begin{equation}
K(x_i,x_j) = \sigma^2_f\exp[(-\frac{1}{2})\sum_{m=1}^{D}\frac{(x_{im}-x_{jm})^2}{ \sigma^2_m}],
\label{eq:3}
\vspace{-0.08in}
\end{equation}
\end{small}\noindent
where $\sigma^2_f$ is the hyper-parameter to be optimized and $\sigma^2_m$ is the covariance of the $m^{th}$ dimension. Finally, the log-likelihood of the trained model can be expressed as
\vspace{-0.08in}
\begin{small}
\begin{equation}
\begin{split}
\log P(E(X)|X^{1:D}) &= -\frac{1}{2}E(X)^T(K+\sigma^2_DI)^{-1}E(X)\\
&-\frac{1}{2}\log det(K+\sigma^2_DI)-\frac{D}{2}\log2\pi,
\end{split}
\label{eq:log}
\vspace{-0.08in}
\end{equation}
\end{small}\noindent
where $I$ is an identity matrix. EPAM is first trained offline with the observation data points, then is run with an application alongside. The prediction is done either simultaneously or at the end of an application. In this research, we train the model with a dataset containing $85,500$ data, validate with $19,496$, and test with $10,000$ data. 

\vspace{-0.12in}

\section{Experimental Setup} \label{detailsExperiment}
\vspace{-0.08in}

\begin{figure}[hbt!]
\centering
\subfigure[Image classification]
{\includegraphics[width=0.35\textwidth]{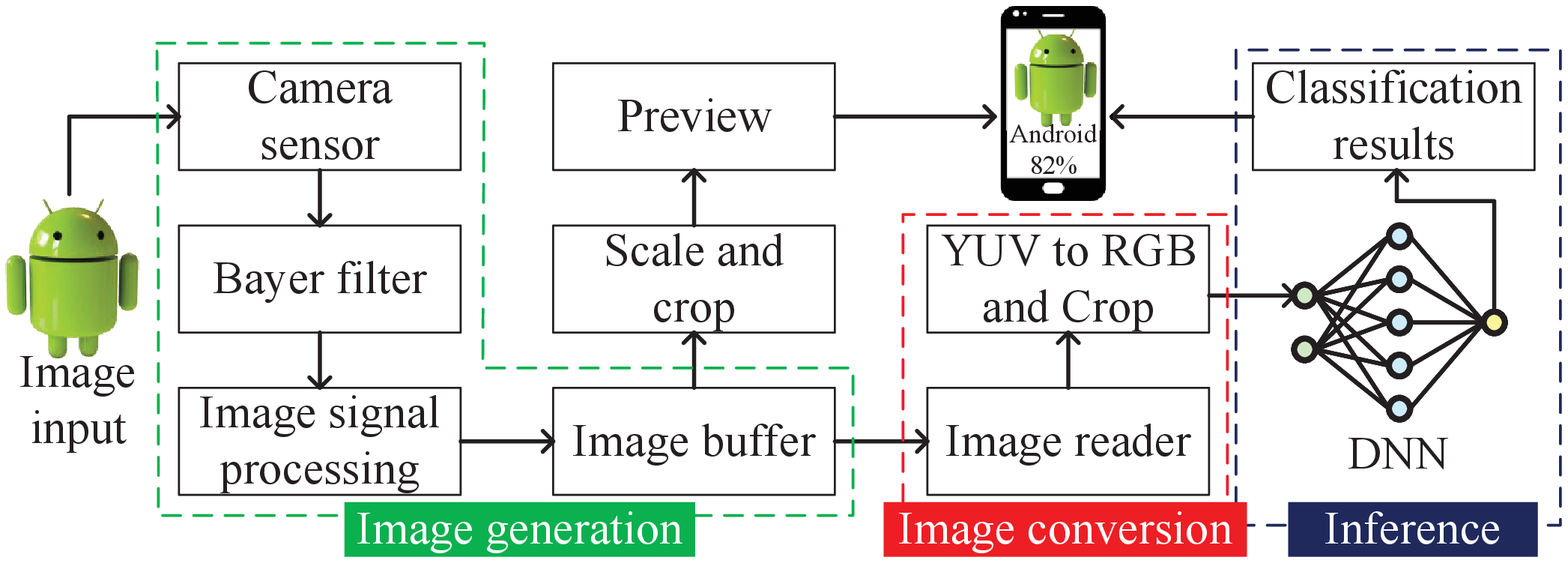}\label{fig:Class}}
\subfigure[Natural language processing (QA)]
{\includegraphics[width=0.35\textwidth]{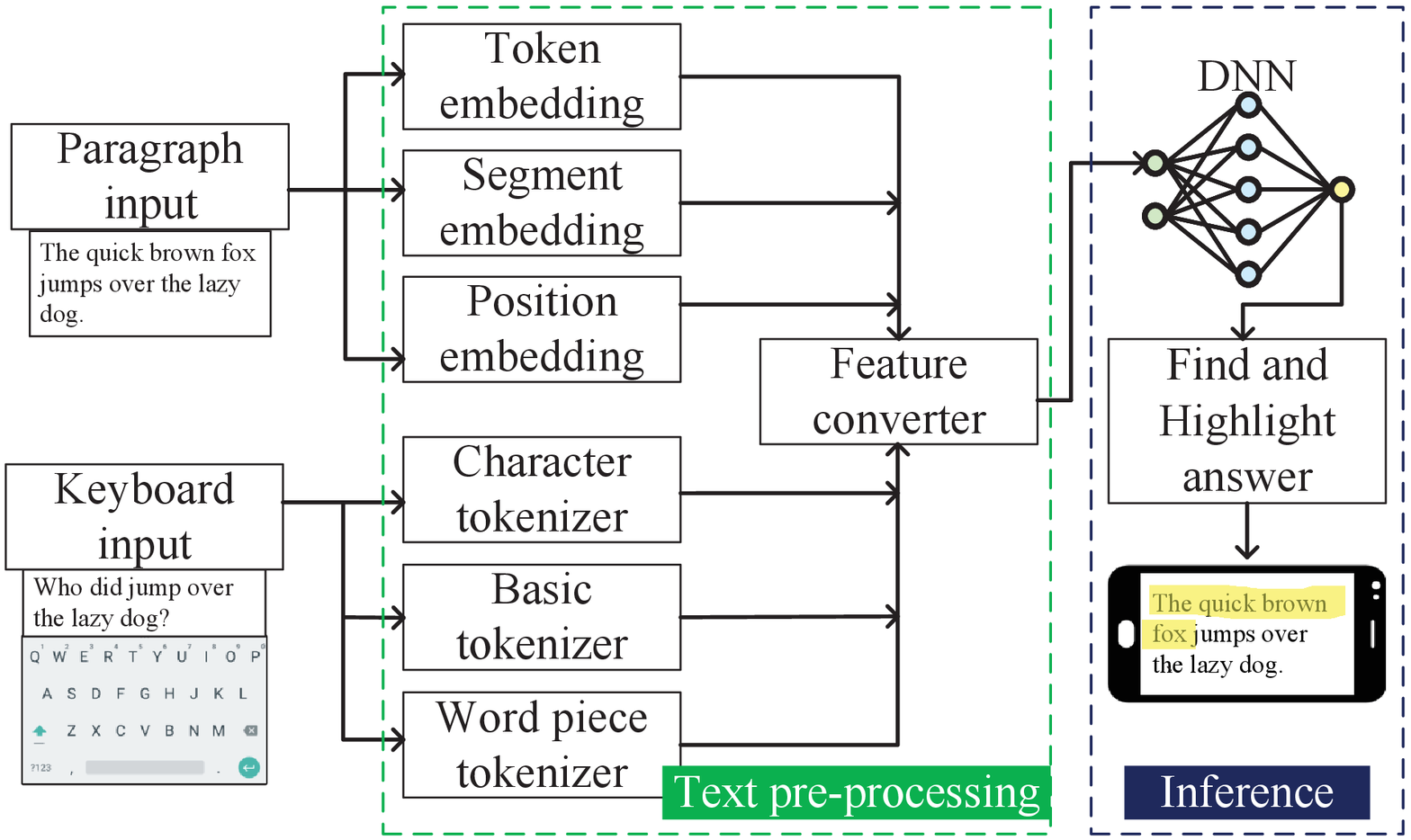}\label{fig:NLP}}
\subfigure[Speech recognition]
{\includegraphics[width=0.35\textwidth]{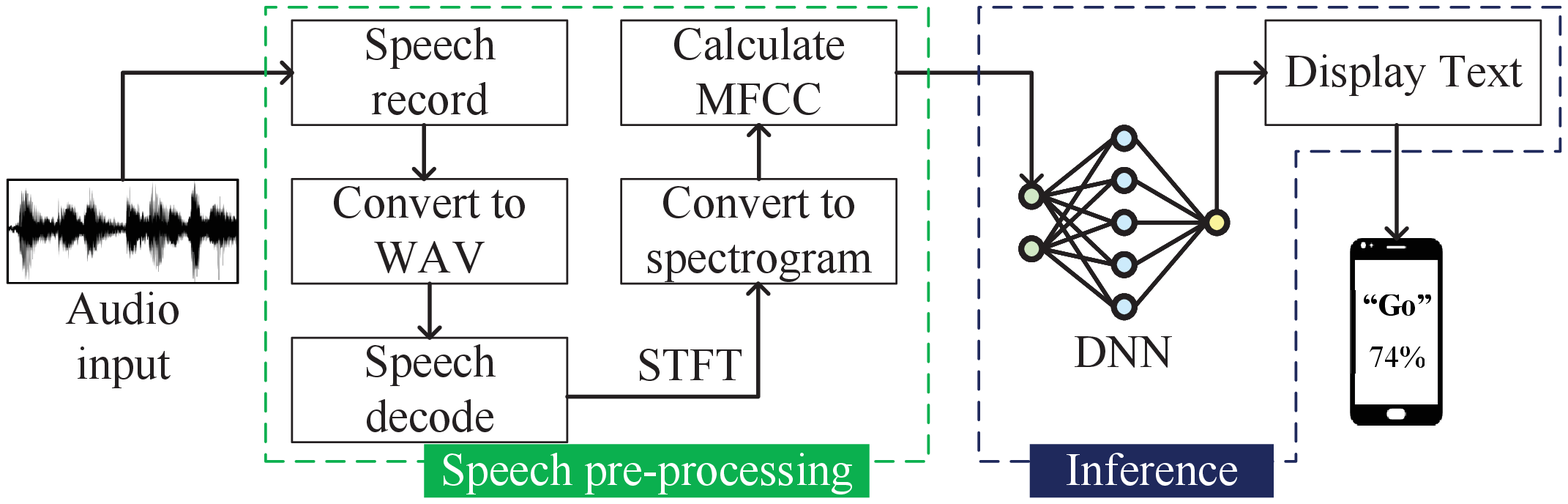}\label{fig:Speech}}
\vspace{-0.1in}

\caption{Pipelines of the mobile AI applications studied in this research.}
\label{fig:Pipelines}   
\vspace{-0.3in}
\end{figure}

\textbf{a) AI applications:} Three mobile AI applications are used in this research: image classification, NLP, and speech recognition. In image classification, as shown in Fig. \ref{fig:Class}, first, the image is captured by the camera sensor, which then goes through a Bayer filter and image signal processor, and, then is stored in an image buffer. The image frame is then scaled and cropped to be previewed while simultaneously going to an image reader, converted from YUV color format to RGB, and cropped according to the input size of the DNN model. Then the converted and cropped frame is taken as the DNN input, generating the classification results to display. 

\par The NLP question-answer application takes both the paragraph input and the question input from the keyboard (Fig. \ref{fig:NLP}. The paragraph is then represented with token, segment, and position embeddings. The keyboard input goes through character, basic, and word piece tokenizer. These embeddings and tokens are passed to a feature converter providing input to the DNN model. The model finds the answer to the question input and highlights it in the paragraph. 

\par Speech recognition application records, converts, and decodes the audio input. The decoded audio signal is converted to a spectrogram by running a short-time Fourier transform (STFT) along with the calculation of the Mel frequency cepstral coefficients (MFCCs). The spectrogram and MFCC are passed to the DNN model. The predicted word is then displayed on the phone as depicted in Fig. \ref{fig:Speech}.

\textbf{b) Testbed:} We implement the applications mentioned above on four Android OS-based smartphones from different manufacturers with distinct configurations to make the measurement study robust with a wide range of parameters. Table \ref{table:phone specs} shows the specifications of the smartphones used in the experiment. However, the intended thorough investigation of mobile AI brings several challenges during the experiment.

\begin{table*}[hbt]
\setlength{\tabcolsep}{5pt}
\caption{Brief specifications of the devices used in the experiments}
\vspace{-0.15in}
\begin{center}

\begin{tabular}{c||ccccccccc}
\hline
\textbf{Denotation}   & \textbf{Model} & \textbf{SoC} & \textbf{CPU}          & \textbf{GPU}         & \textbf{Dedicated AI} & \textbf{RAM} & \textbf{OS}          & \textbf{NNAPI}       & \textbf{Release} \\
\textbf{}             & \textbf{}      & \textbf{}    & \textbf{}             & \textbf{}            & \textbf{accelerator}  & \textbf{}    & \textbf{}            & \textbf{support}     & \textbf{Date}    \\ \hline  

Device-1 & Huawei & Kirin 9000 & 8-core (1x3.13GHz A77 & Mali G78 & Ascend Lite+ & 8GB & Android 10 & Yes & October,\\
& Mate & (5 nm) & 3x2.54GHz A77 & & Tiny NPU & LPDDR5 & & & 2020\\
& 40 Pro & & 4x2.05GHz A55) & & Da Vinci 2.0 & & & &\\

Device-2 & OnePlus & Snapdragon & 8-core (1x2.84GHz & Adreno 650 & Hexagon & 8GB & Android 10 & Yes & April,\\
& 8 Pro & 865 (7 nm) & 3x2.42GHz & & 698 DSP & LPDDR5 & & & 2020\\
& & & 4x1.8GHz Kryo 585) & & & & & & \\

Device-3 & Motorola & Helio P70 & 8-core (4x2.0GHz A73 & Mali G72 & MediaTek & 4GB & Android 9 & Yes & October,\\
& One Macro & (12 nm) & 4x2.0GHz A53) & & APU & LPDDR4X & & & 2019\\

Device-4 & Xiaomi & Snapdragon & 8-core (4x2GHz Gold & Adreno 610 & Hexagon & 4GB & Android 10 & Yes & August,\\
& Redmi & 665 (11 nm) & 4x1.8GHz Silver & & 686 DSP & LPDDR4X & & & 2020\\
& Note8 & & Kryo260) \\

\hline
\end{tabular}
\label{table:phone specs}
\end{center}
\vspace{-0.25in}
\end{table*}

\begin{table*}[htb!]
\caption{DNN models used in this research}
\centering
\setlength{\tabcolsep}{10pt} 
\vspace{-0.1in}
\begin{tabular}{c||c c c c c}
\hline
\textbf{Denotation} & \textbf{Model Name}             & \textbf{Application}          & \textbf{Input size} & \textbf{No. of layers} & \textbf{Model Size} \\
\hline
Model 1 & MobileNetV1 (Float) & Image classification & 224x224x3 & 31 & 16.9 MB\\
Model 2 & MobileNetV1 (Quantized) & Image classification & 224x224x3 & 31 & 4.3 MB\\
Model 3 & EfficientNet-lite (Float) & Image classification & 224x224x3 & 62 & 18.6 MB\\
Model 4 & EfficientNet-lite (Quantized) & Image classification & 224x224x3 & 65 & 5.4 MB\\
Model 5 & NASNet Mobile (Float) & Image classification & 224x224x3 & 663 & 21.4 MB\\
Model 6 & Mobile BERT QA & Natural language processing & int32 {[}1, 384{]} & 2541 & 100.7 MB\\
Model 7 & Tensorflow ASR & Speech recognition & {[}20 Hz, 4 kHz{]}  & 8 & 3.8 MB\\ 
\hline
\end{tabular}
\label{table:DNN}
\vspace{-0.25in}
\end{table*}


\par Android Studio, along with other third-party contributors, provides developers with memory and battery profilers, which cannot generate the data necessary to measure memory usage and power consumption precisely. In this experiment, we collect latency timestamp data of each segment of a mobile AI pipeline along with their corresponding memory usage. To measure the energy consumption, we use an external power measurement tool ``Monsoon Power Monitor'' that provides data sampled at every 0.2 ms interval. However, due to the delicate design of power input terminals, the latest smartphones need to be heated and opened to remove the battery, and then are connected to the power monitor. After careful measurement of power data, they are matched with the corresponding latency timestamps.


\par To make the experiment environment controllable, we carry out all the experiments in a similar condition, e.g., brightness, camera focus, image resolution, background applications, processing sources, and test dataset. We use $640\times480$ pixels as the image resolution, and \hytt{TensorFlow Lite Delegate} to control the processing sources. The 2017 COCO test dataset, WH-questions, and fixed single words are used for testing the classification, NLP, and speech recognition, respectively. In addition, even without any applications running in the background, there is always a minimal power consumption -- which we call the \textit{base power}. To distinguish the mobile AI power from the base power, an additional layer is used before the actual AI application.




\textbf{c) AI models:} In this research, we use seven DNN models for three different applications. In Table \ref{table:DNN}, the details of each model, including the input size, number of layers, and the trained model size (occupied storage space) are shown.

\textbf{d) Performance metric:} We evaluate all the AI applications' performances in terms of their latency, energy consumption, and memory usage. The total energy consumption is controlled by latency and memory usage, as well as the category of AI applications, processing sources, model types (float and quantized), and DNN structure and model size.

\vspace{-0.1in}
\section{Results and Discussion}\label{sec:results}

\vspace{-0.1in}
We conduct experiments with all the devices listed in Table \ref{table:phone specs} and models listed in Table \ref{table:DNN} by switching to different processing sources, such as CPU thread 1 and thread 4, GPU, and NNAPI. Models 1 to 5 are for vision-based AI, and models 6 and 7 are for non-vision-based AI applications. It is to be noted that models 2, 4, 6, and 7 do not support GPU processing due to a lack of \hytt{TensorFlow Lite} optimization. In general, the applications have input data processing (combining image generation and conversion in classification) and inference tasks. In this paper, we show some of the interesting findings due to space constraints.

\vspace{-0.1in}
\subsection{Latency and energy consumption of mobile AI}\label{subsec:latencyEnergyAll}
\vspace{-0.1in}

\begin{figure}[t]
\centering
\subfigure[]
{\includegraphics[width=0.235\textwidth]{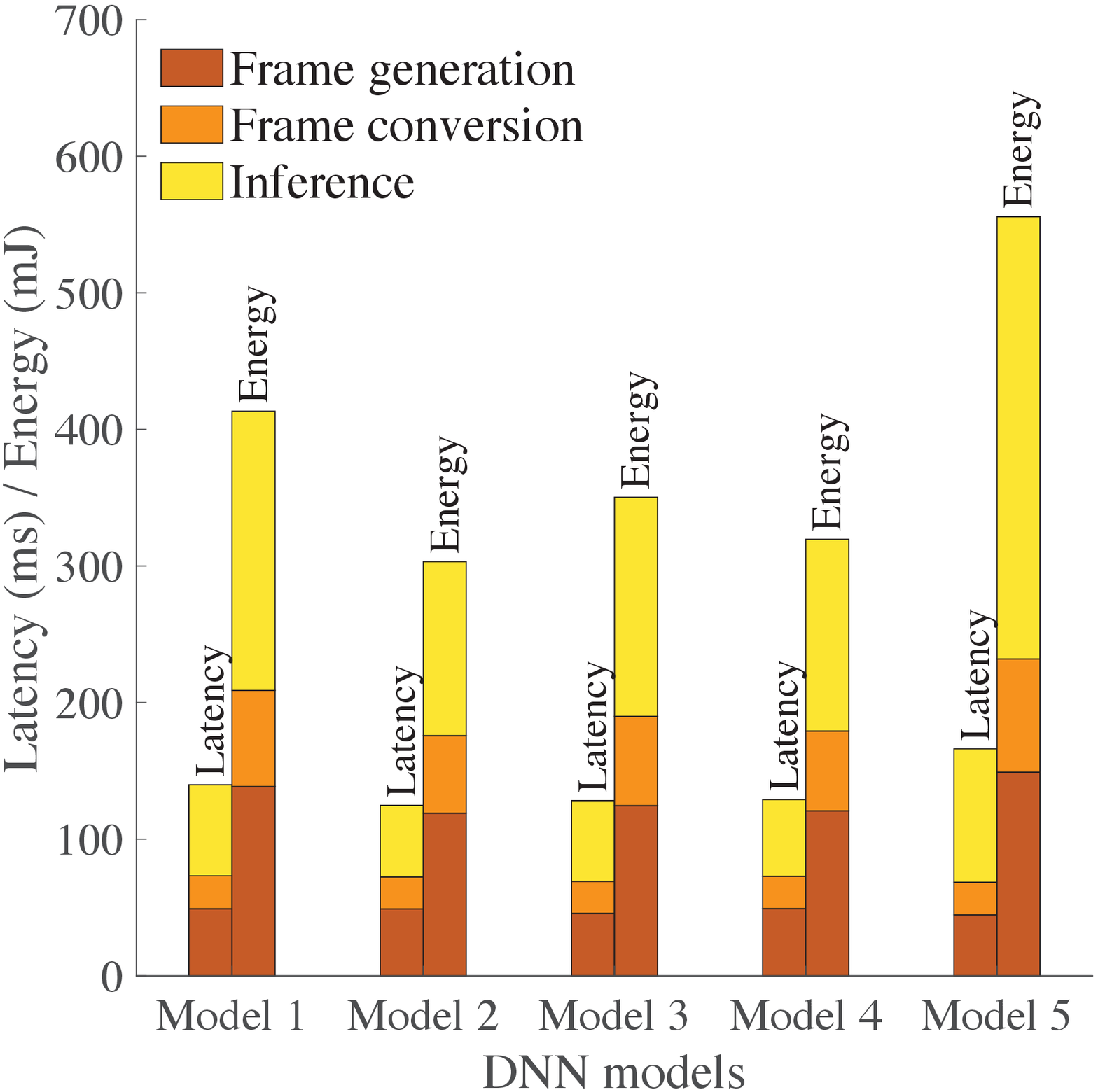}\label{fig:LatEnerCPU1}}
\hspace*{\fill}
\subfigure[]
{\includegraphics[width=0.235\textwidth]{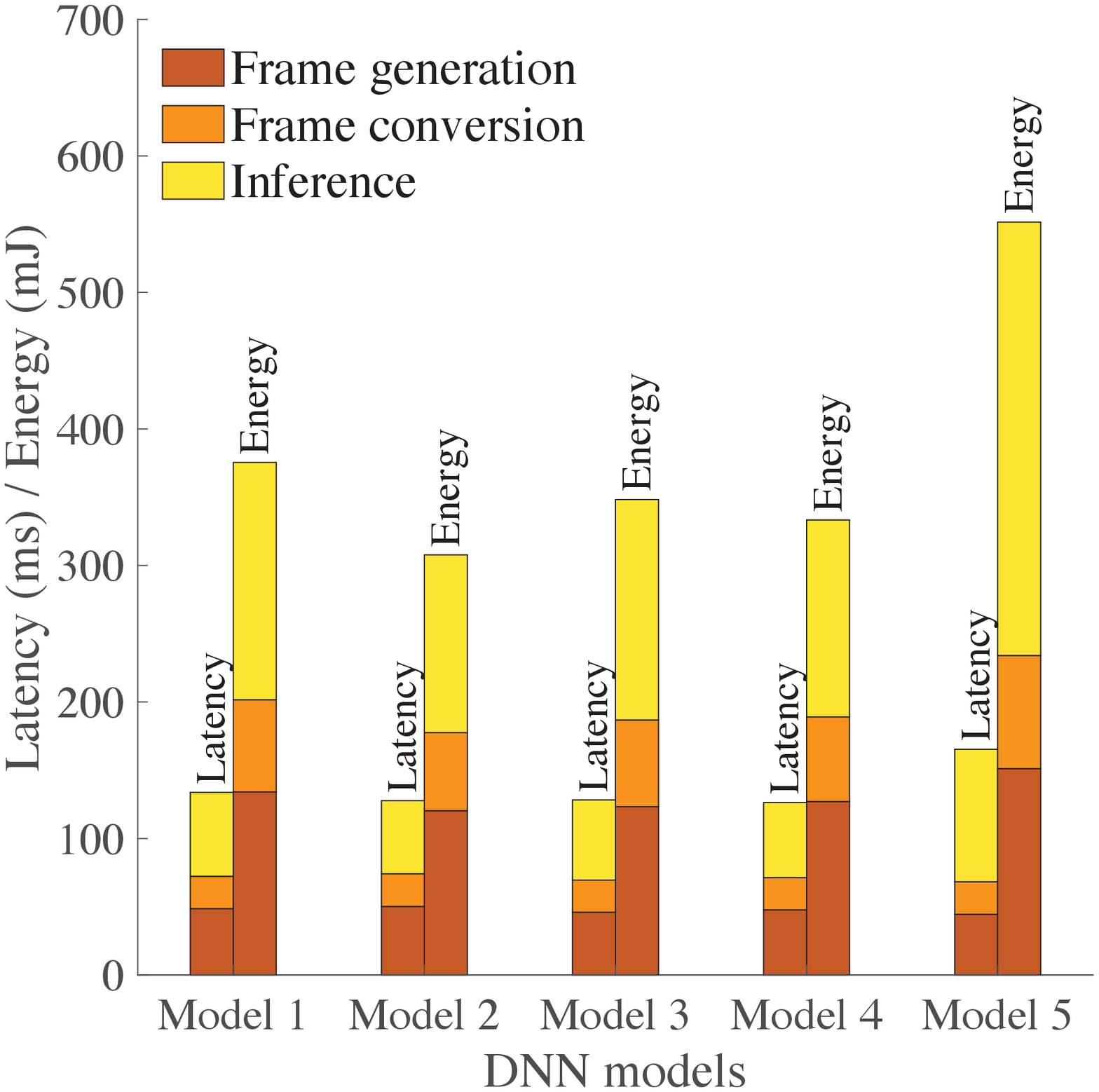}\label{fig:LatEnerCPU4}}
\vspace{-0.15in}

\subfigure[]
{\includegraphics[width=0.235\textwidth]{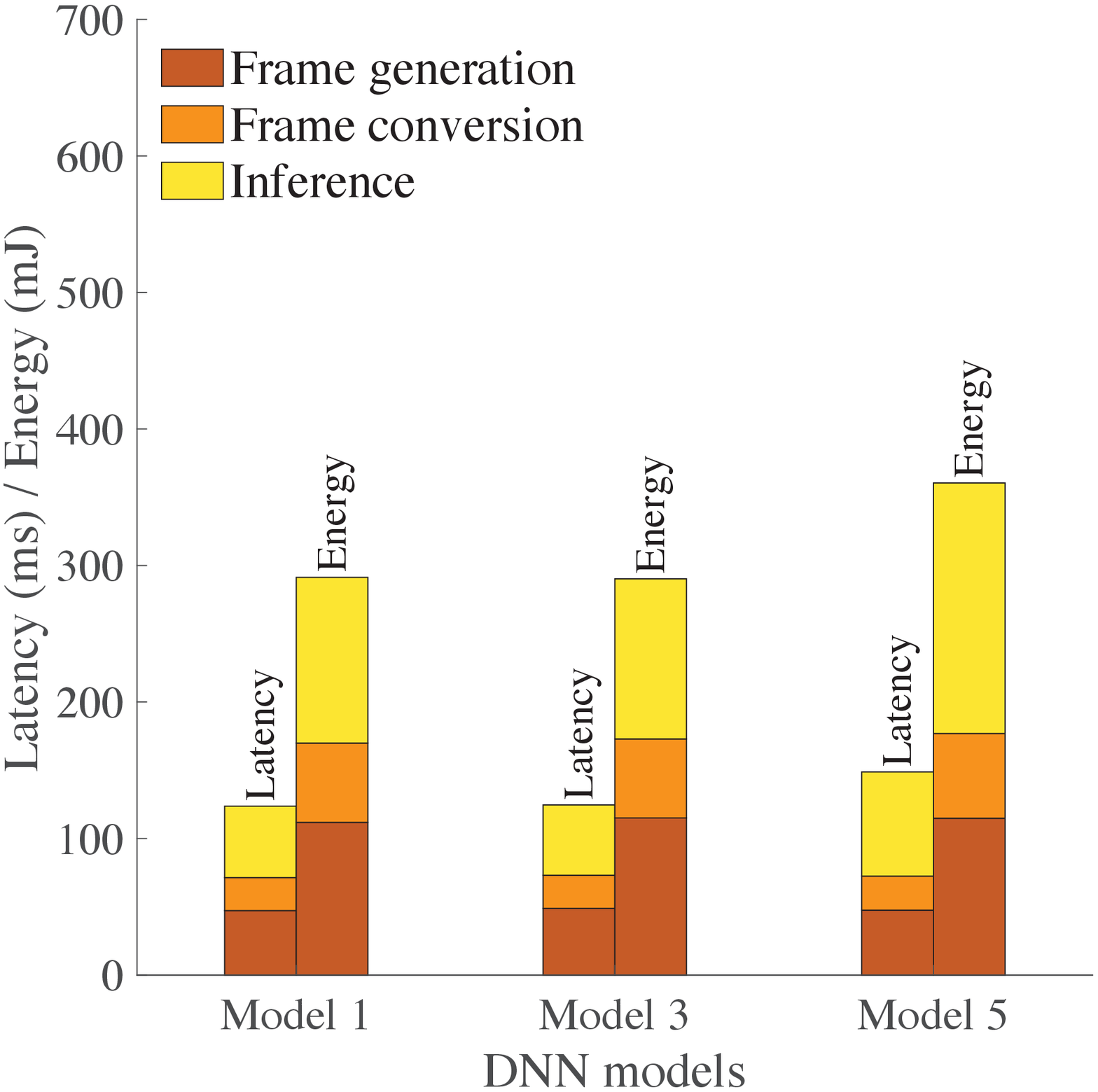}\label{fig:LatEnerGPU}}
\hspace*{\fill}
\subfigure[]
{\includegraphics[width=0.23\textwidth]{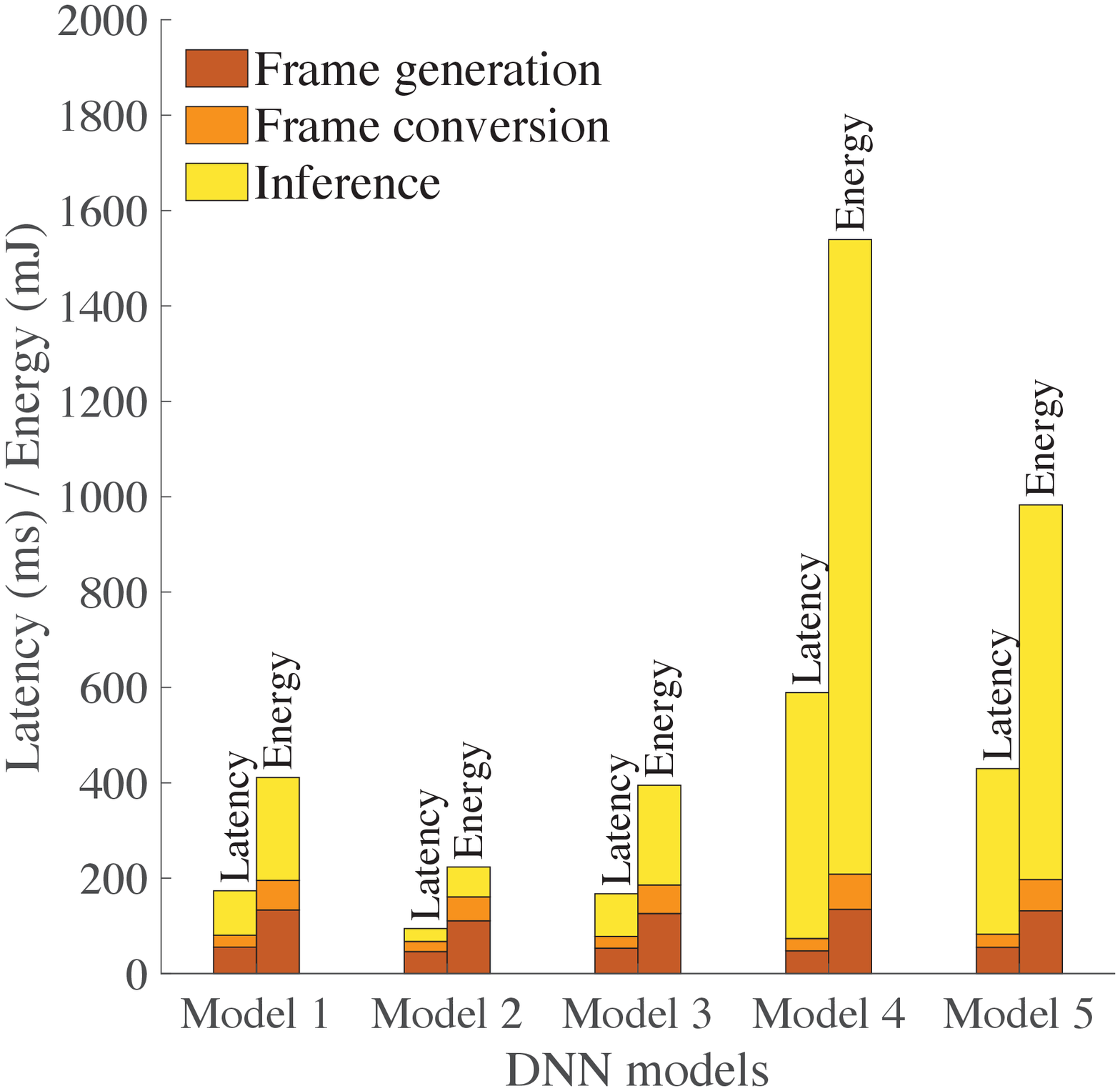}\label{fig:LatEnerNNAPI}}
\vspace{-0.15in}

\subfigure[]
{\includegraphics[width=0.23\textwidth]{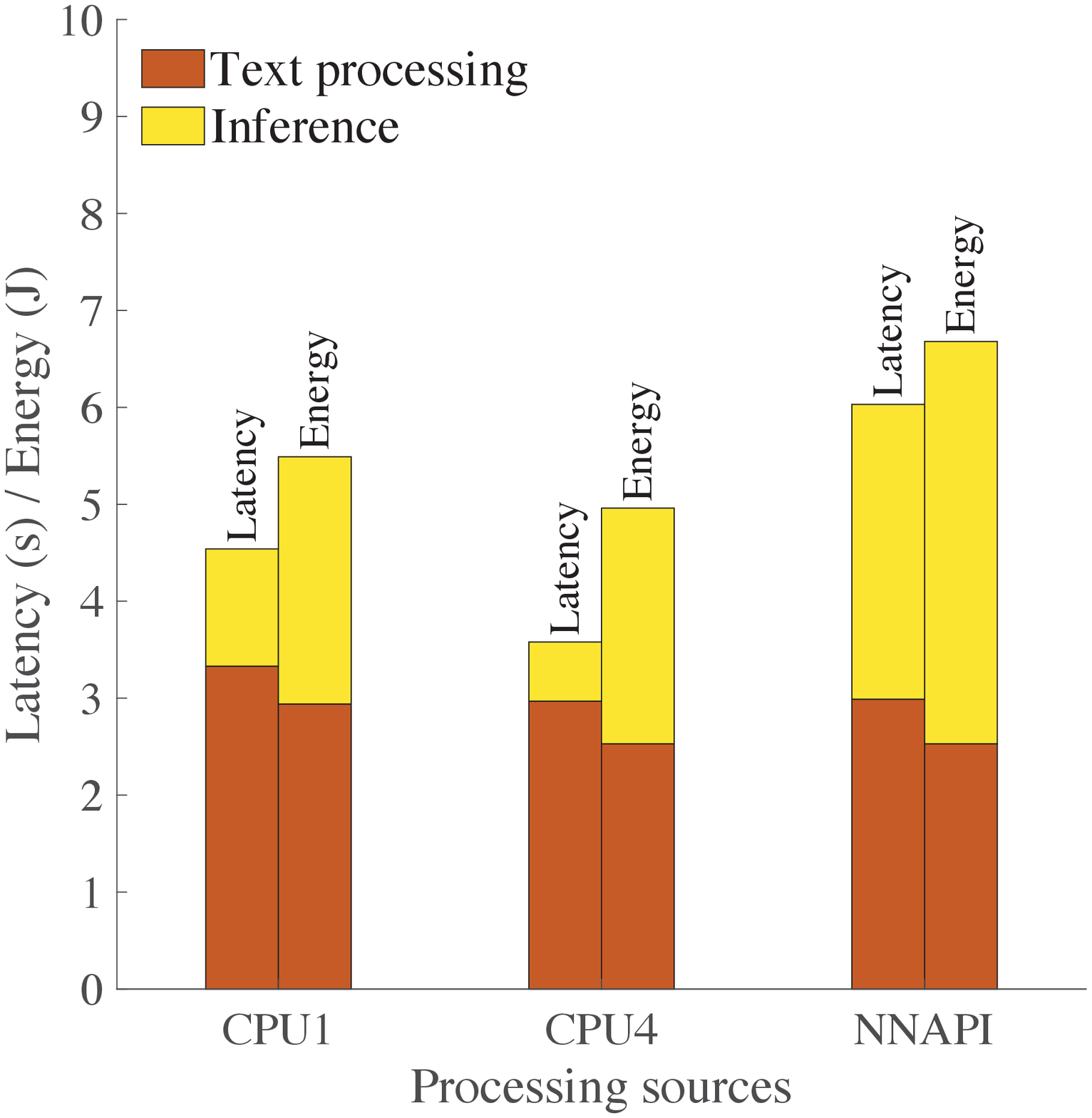}\label{fig:LatEnerNLP}}
\hspace*{\fill}
\subfigure[]
{\includegraphics[width=0.23\textwidth]{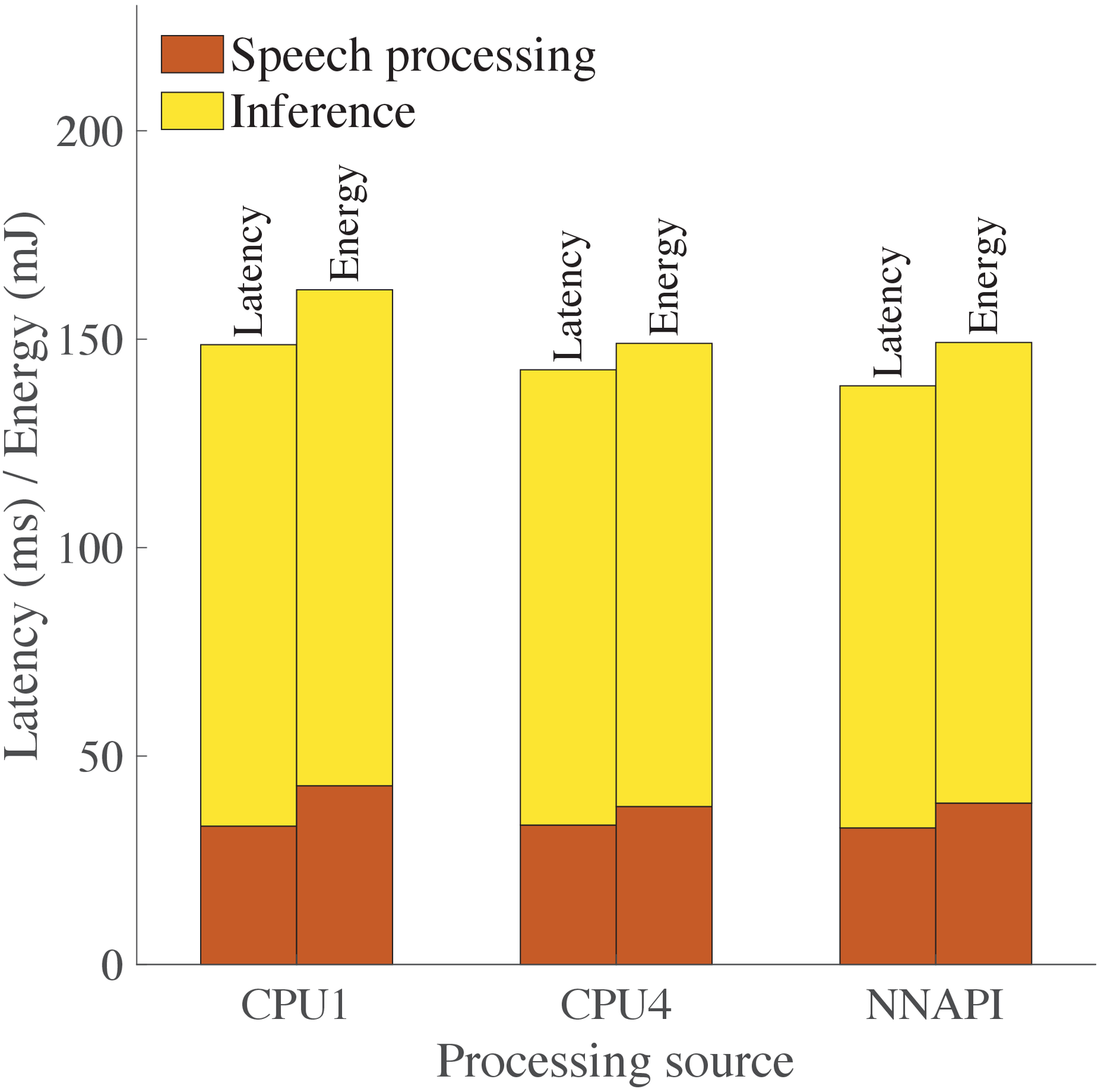}\label{fig:LatEnerSpeech}}
\vspace{-0.15in}
\caption{End-to-end mean latency and energy consumption per cycle of vision-based models 1--5 for (a) single- and (b) multi-thread CPU, (c) GPU, and (d) NNAPI, and non-vision-based (e) model 6 and (f) model 7.}
\label{fig:LatEnerAll}   
\vspace{-0.25in}
\end{figure}

The end-to-end latency and energy consumption per cycle for all the models with different processing sources are shown in Fig. \ref{fig:LatEnerAll}. First, we can see that quantized models decrease the inference latency ($13\%$) and energy consumption ($25\%$) from their respective float models. Additionally, there is a reduction in the overall latency of $4\%$ when switching to a 4-thread from a single-thread CPU. However, in quantized models, the multi-thread CPU processing slightly increases the total energy consumption ($3\%$ on average). The use of GPU even lowers the end-to-end latency and energy consumption compared to the use of single-thread CPU ($8\%$ and $27\%$ respectively on average) and 4-thread CPU ($7\%$ and $25\%$ respectively on average). On the contrary, NNAPI behaves differently than the other three processing sources on different devices. For models 4 and 5, NNAPI increases latency and energy considerably. Our insight here is that NNAPI can perform better with sufficient hardware support from the manufacturers. 



\vspace{-0.02 in}
\par An interesting fact about the NLP application is that the text processing step shows an entirely different latency pattern. This segment takes user input which does not take uniform time, i.e., it varies with user habits of typing and thinking of the question. Hence, the processing stage here is completely unpredictable for different users. In NLP, each input consumes around $5.7$ J, whereas, another non-vision application, speech recognition takes around $161.85$ mJ to process one speech input sampled at a rate of $16$ kHz using a single-thread CPU. NNAPI consumes the least latency and energy for speech recognition.

\par
\begin{figure}[t]
\centering
\subfigure[]
{\includegraphics[width=0.4\textwidth]{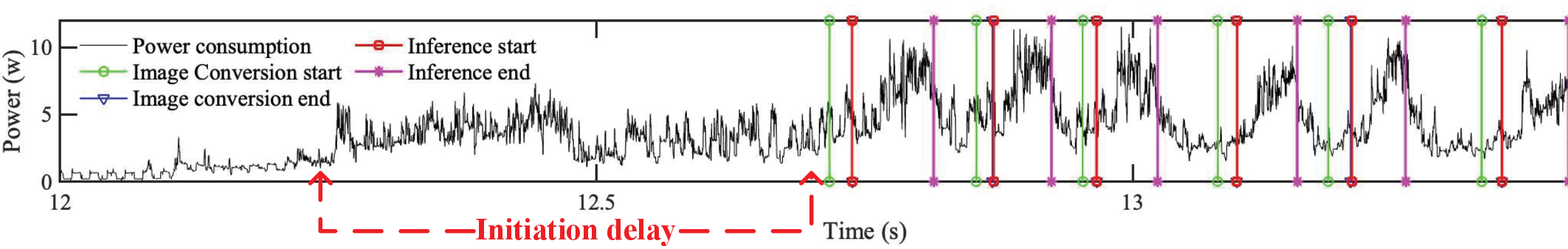}\label{fig:Class_pwr}}\\
\vspace{-0.05in}
\subfigure[]
{\includegraphics[width=0.4\textwidth]{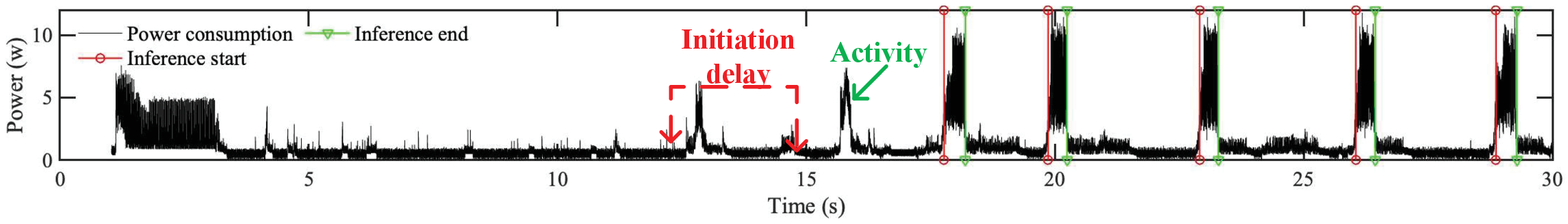}\label{fig:NLP_pwr}}\\
\vspace{-0.05in}
\subfigure[]
{\includegraphics[width=0.4\textwidth]{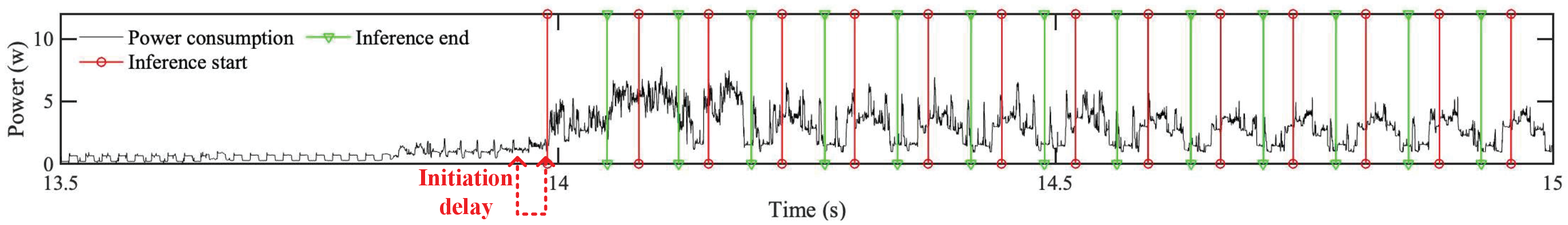}\label{fig:Speech_pwr}}
\vspace{-0.05in}
\caption{Power consumption pattern for (a) classification, (b) NLP, and (c) speech recognition.}
\label{fig:PowerApp}   
\vspace{-0.3in}
\end{figure}

 In addition, we examine the power consumption charts of different applications and processing sources (Fig. \ref{fig:PowerApp}). We observe a slight initiation delay for every application (marked with red arrows in Fig. \ref{fig:PowerApp}), which varies with using different processors and applications. This delay occurs during the time when the application interface initiates till the activity-start point, which is mainly originated by different hardware components being accessed at the beginning of an AI application, such as the camera, keyboard, speaker, and microphone. 
Besides, different processor delegations (e.g., GPU and NNAPI) are also done during this period. 

\textbf{Highlights:} \textit{Non-vision applications cannot be generalized for latency and energy like vision-based ones. GPU processing is not supported by non-vision applications, which should be explored widely. The initiation delay (i.e., the delay between the activity trigger and start point) varies along AI models, processing sources, and applications, which is caused by accessing different hardware components by mobile AI applications.}

%

%




\vspace{-0.1in}
\subsection{DNN structures and their inference latency and energy}\label{subsec:DNN}
\begin{figure}[hbt]
\vspace{-0.15 in}
\centerline{\includegraphics[scale=0.4]{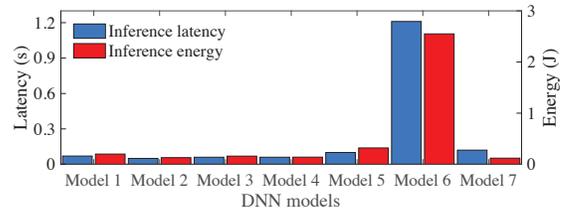}}
\vspace{-0.15in}
\caption{Inference latency and energy consumption per cycle by DNN models.}
\label{fig: Model_Lat_Ener}
\vspace{-0.1in}
\end{figure}

\vspace{-0.05in}
DNN structures define the way inference activities work in a mobile AI application. The behavior of DNN structures varies across different kinds of applications as well, e.g., vision and non-vision AI. For instance, a smaller DNN structure for vision applications can incur higher latency and energy than a larger non-vision DNN structure. Inference latency and energy consumption per cycle are shown in Fig. \ref{fig: Model_Lat_Ener} for DNN models with single-thread CPU processing. We observe that model 5 takes longer inference time and energy due to its larger structure than the other vision-based AI models. The longest latency and highest energy are evident in model 6 (a complex structure comprising 2541 layers). 


\textbf{Highlights:} \textit{DNN structures influence inference latency and energy significantly, but the relationship is not linear at all. Generally, larger DNN structures are responsible for higher latency and energy for a mobile AI application.}

\vspace{-0.1in}
\subsection{DNN model size, memory usage, and inference energy}\label{subsec:modelMem}
\vspace{-0.05in}

\begin{figure}[hbt!]
\vspace{-0.15 in}
\centerline{\includegraphics[scale=0.4]{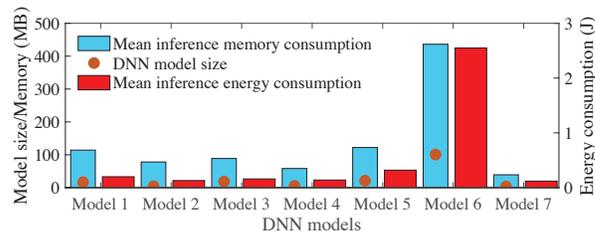}}
\vspace{-0.15in}
\caption{Comparison of DNN model size, inference memory usage, and inference energy consumption.}
\label{fig: Model_size_memory}
\vspace{-0.15 in}
\end{figure}

DNN model size (i.e., the storage space occupied by the model) impacts memory usage and energy consumption during inference. From our experiment, we observe that model 7 has the lowest model size, hence causing the lowest memory and energy consumption, whereas model 6 has the highest size, memory, and energy consumption. This is more evident from Fig. \ref{fig: Model_size_memory}, which shows a comparison among all the models' sizes, inference memory, and energy consumption for single-thread CPU processing.


\textbf{Highlights:} \textit{Lower memory used by mobile AI applications ensures computation resources and energy for other mobile device activities. From this perspective, quantized and smaller DNN models are best suited for mobile AI. The larger the storage occupied by a DNN model, the higher the memory and energy consumption.}

\subsection{Performance evaluation of EPAM}\label{subsec:performanceEPAM}
\begin{figure}[hbt]
\vspace{-0.15in}
\centering
\subfigure[]
{\includegraphics[width=0.23\textwidth]{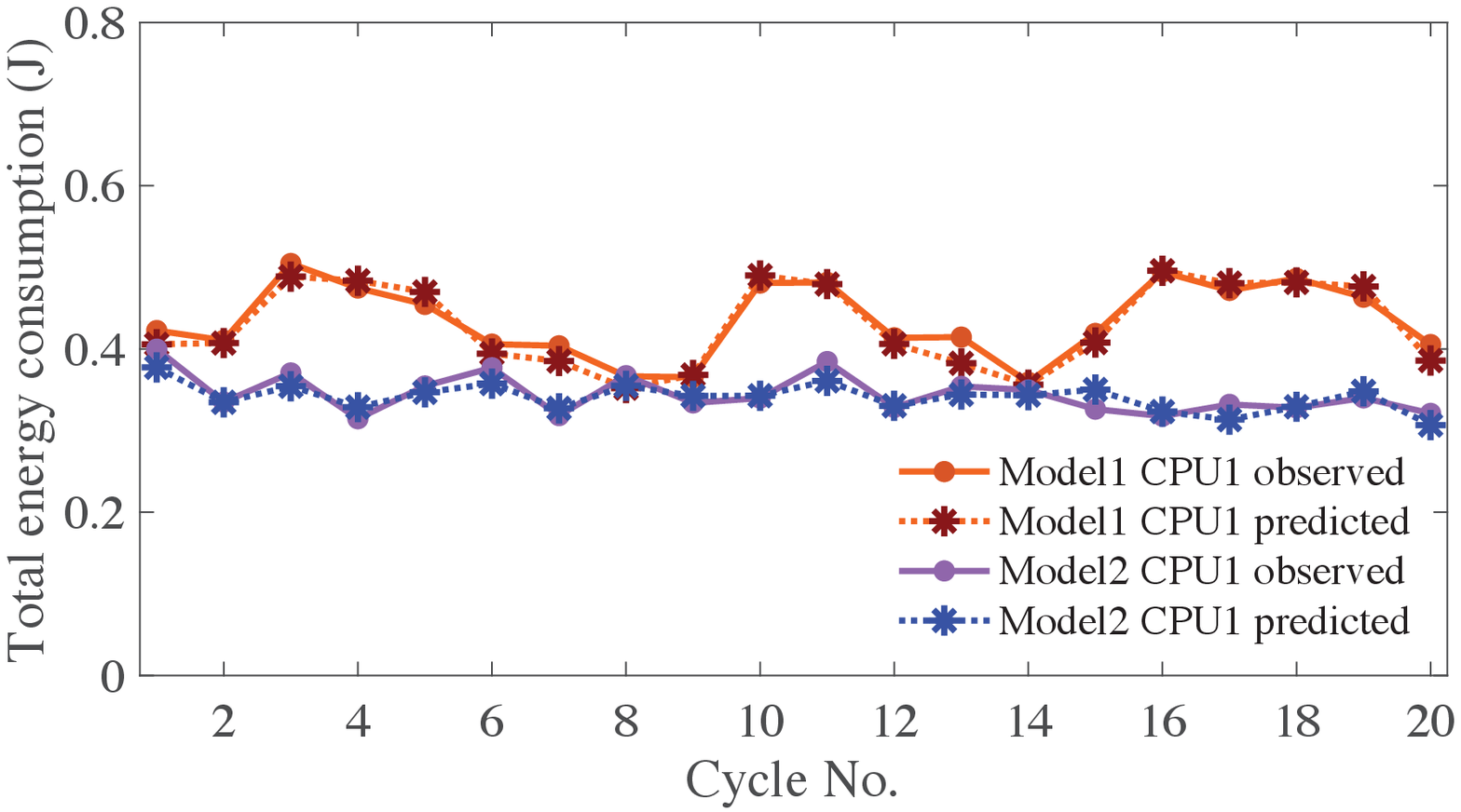}\label{fig:EPAM_mobileNet}}
\hspace*{\fill}
\subfigure[]
{\includegraphics[width=0.235\textwidth]{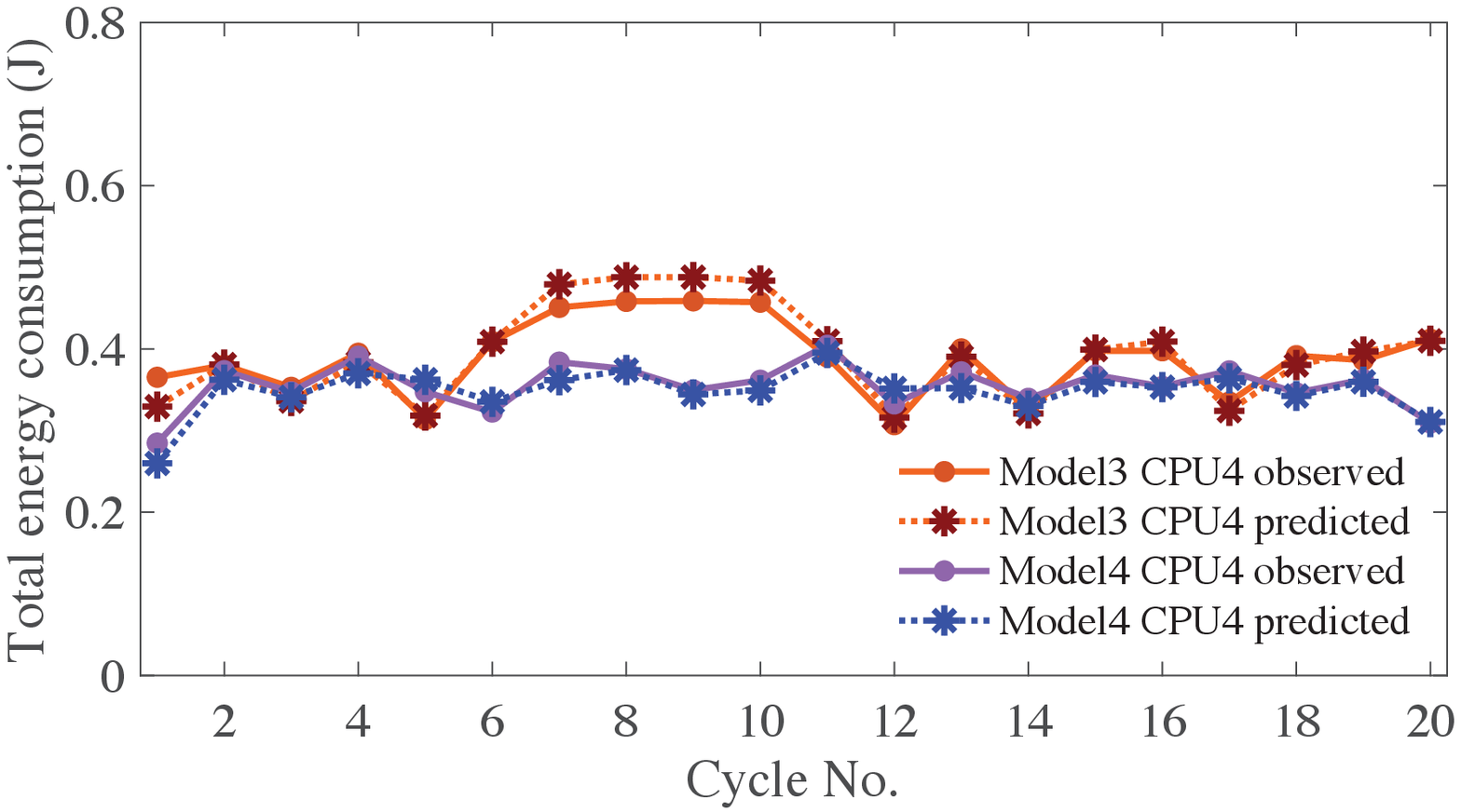}\label{fig:EPAM_effNet}}
\vspace{-0.15in}

\subfigure[]
{\includegraphics[width=0.225\textwidth]{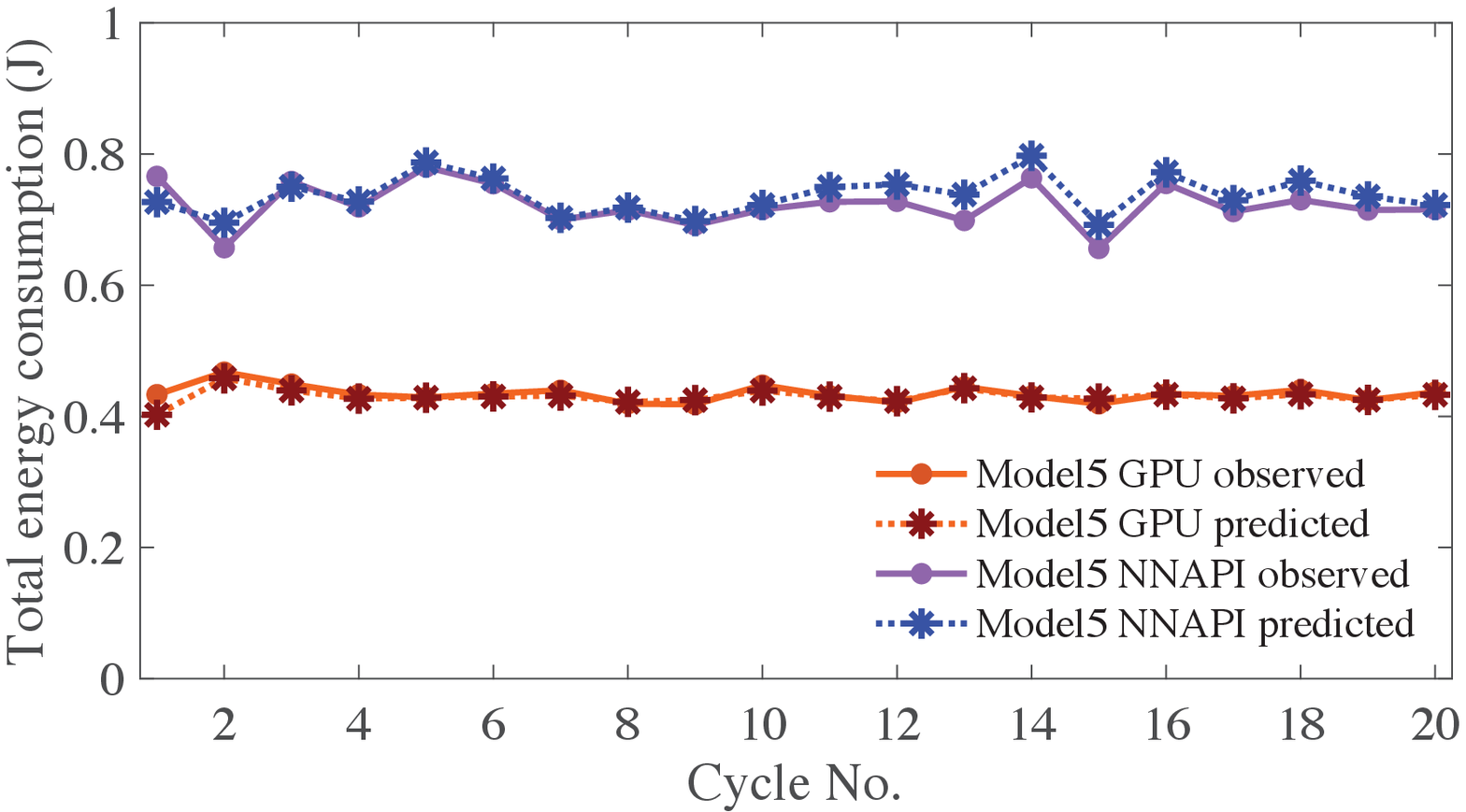}\label{fig:EPAM_nasNet}}
\hspace*{\fill}
\subfigure[]
{\includegraphics[width=0.238\textwidth]{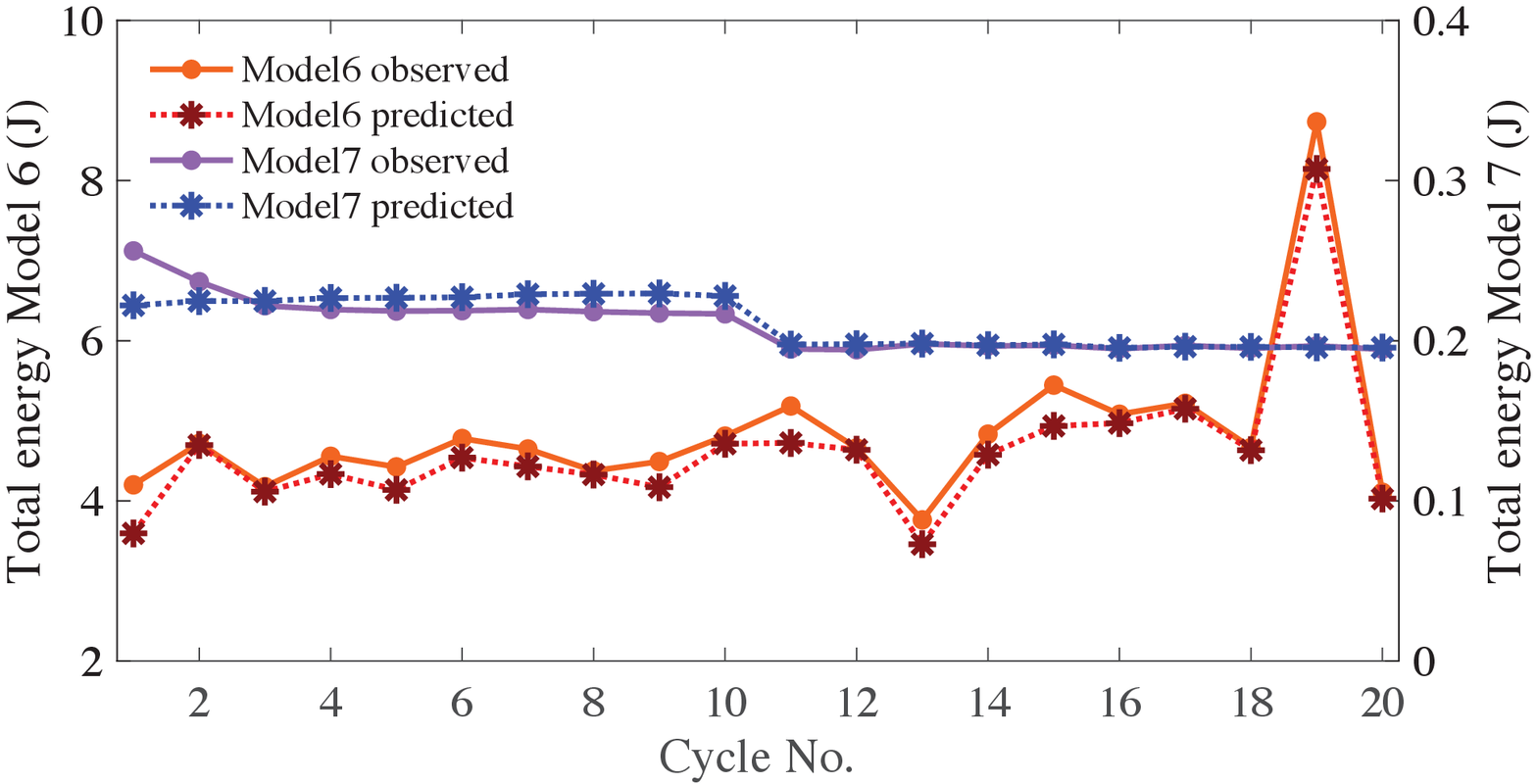}\label{fig:EPAM_asrBert}}
\vspace{-0.15in}

\caption{Evaluation of EPAM for (a) model 1 and 2 (b) model 3 and 4, (c) model 5, and (d) model 6 and 7 with different processing sources.}
\label{fig:EPAMevaluation}   
\vspace{-0.1in}
\end{figure}

\vspace{-0.05in}
We develop and train the Gaussian process regression-based predictive energy model, EPAM, with each device's SoC, CPU frequency, no. of cores, memory size, processing sources, no. of threads, application type, DNN model, DNN structure, memory usage, processing latency, and inference latency from the large experimental dataset from this research to predict the total energy consumption per application cycle (data processing and inference for each input). We use an empty basis function, and ARD squared exponential kernel function for the hyper-parameter optimization. We use device-1, 2, and 4 for training and validation, and device-3 for 1-step ahead prediction testing. Due to page limitation, we show only a few prediction results in Fig. \ref{fig:EPAMevaluation}. We observe that EPAM's energy prediction per cycle is highly accurate for all the models. The overall root mean squared error (RMSE) is $0.075$ ($3.06\%$), and the marginal log-likelihood value is $-1.449\times10^2$, which show that the trained model is a good fit for the prediction. The prediction latency depends on the machine used in running the model. 


\textbf{Highlights:} \textit{EPAM further helps developers and users to perceive the performance of individual AI applications in terms of energy with high accuracy -- which is the primary motivation of this research work. The larger and more diverse the training dataset, the higher the prediction accuracy.}

\vspace{-0.1in}
\section{Conclusion}
\vspace{-0.08in}
In this paper, we presented a comprehensive study of mobile AI applications with different processing sources and AI models. Overcoming the challenges with measurement, we conducted experiments to assess the performance of different AI models, processing sources, and devices. Our measurement work shows that the latency, energy consumption, and memory usage vary based on DNN models and processing sources. Mobile AI systems' performance is substantially improved using quantized models than floating-point models in terms of latency and energy. Another important finding is that the storage space occupied by DNN models influences the memory and energy consumed during inference almost linearly. Additionally, non-vision applications follow a different trend of latency and energy consumption than vision-based AI since their input processing techniques differ from vision applications. Every AI application has an initiation delay caused by accessing various hardware components of mobile devices, which varies for different models and configurations. Moreover, the latency, memory, AI model, and device configuration impact the total energy consumption for a complete application cycle, albeit at different correlations. This non-linear correlation in a non-parametric model led to our proposed predictive energy model, EPAM, based on Gaussian process regression. Finally, we trained and validated EPAM with the vast dataset obtained from our experiment. The evaluation of EPAM shows high accuracy with an overall RMSE of 0.075 ($3.06\%$). Developers can use EPAM to predict the energy consumption of their mobile AI applications without measuring the energy externally to improve the comprehensive user experience. To summarize, this novel predictive energy model, EPAM, will help the mobile AI research community design energy-improved applications considering all the control factors and parameters that can reduce energy requirements to enable better service for smartphones, wearable devices, and autonomous vehicles.


\vspace{-0.05in}
\linespread{0.89}

\end{document}